\documentclass[11pt, a4paper, logo]{omni}

\usepackage[utf8]{inputenc}
\usepackage[T1]{fontenc}
\usepackage{url}
\usepackage{graphicx}
\usepackage{booktabs}
\usepackage{amsmath,amssymb,mathtools}
\usepackage{amsfonts}
\usepackage{microtype}
\usepackage{xcolor}
\usepackage{multirow}
\usepackage{colortbl}
\usepackage{array}
\usepackage{xspace}
\usepackage{cite}
\usepackage{marvosym}

\usepackage{amsmath,amsfonts,bm}

\def\eqref#1{equation~\ref{#1}}

\def\1{\bm{1}}

\DeclareMathAlphabet{\mathsfit}{\encodingdefault}{\sfdefault}{m}{sl}
\SetMathAlphabet{\mathsfit}{bold}{\encodingdefault}{\sfdefault}{bx}{n}

\definecolor{lightblue}{HTML}{0071bc}
\usepackage[pagebackref=false, breaklinks=true, colorlinks, citecolor=lightblue, urlcolor=lightblue, linkcolor=lightblue, bookmarks=false]{hyperref}
\usepackage[capitalize,noabbrev]{cleveref}

\newcommand{\name}{\texttt{Omni-Sleep}\xspace}
\newcommand{\equalcontrib}{\textsuperscript{*}}
\newcommand{\corrauth}{\textsuperscript{\Letter}}

\title{Omni-Sleep: A Sleep Foundation Model via Hierarchical Contrastive Learning of CNS--ANS Dynamics}

\author[1,3]{Zhoujie Hou\equalcontrib}
\author[1,3]{Song Wang\equalcontrib}
\author[1,3]{Kexin Lou\equalcontrib}
\author[1,3]{Mo Wang}
\author[1,3]{Chen Wei}
\author[1,2,3]{Quanying Liu\corrauth}

\affil[1]{Department of Biomedical Engineering, Southern University of Science and Technology, Shenzhen, China}
\affil[2]{Shenzhen Loop Area Institute, Shenzhen, China}
\affil[3]{Omni-Intelligence, Shenzhen, China}

\correspondingauthor{\equalcontrib Equal contribution. \corrauth Correspondence to: liuqy@sustech.edu.cn.}
\codelink{https://github.com/AutoBrain-sleep/OmniSleep}
\keywords{Sleep; self-supervised learning; multimodal PSG; foundation model}

\begin{abstract}
Sleep physiology arises from the coordinated dynamics of the central nervous system (CNS) and autonomic nervous system (ANS), as reflected by multimodal polysomnography signals including EEG, EOG, EMG, ECG, and respiration. However, existing sleep foundation models often fuse heterogeneous biosignals in a topology-agnostic manner, overlooking their physiological organization. We introduce Omni-Sleep, a sleep foundation model that uses the CNS/ANS partition as a physiological prior for topology-constrained representation learning. Omni-Sleep learns structured representations through three objectives: intra-system consistency, which captures shared subsystem-level factors within neural and cardio-respiratory signals; inter-system synchronization, which aligns subsystem trajectories to model brain--body dynamics; and latent-space masked temporal modeling, which captures long-horizon sleep dynamics. Pre-trained on over 100,000 hours of multi-center multimodal PSG data, Omni-Sleep is evaluated on sleep staging and multi-disease classification. Across datasets and modality-ablation settings, Omni-Sleep outperforms strong foundation-model baselines, showing improved label efficiency, cross-dataset generalization, and robustness to missing modalities. These results highlight the value of physiological hierarchy for generalizable sleep representation learning. Code is available at \url{https://github.com/AutoBrain-sleep/OmniSleep}.
\end{abstract}

\begin{document}

\maketitle

\section{Introduction}

Sleep is a dynamic physiological process governed by coordinated interactions between central neural activity and autonomic cardio-respiratory regulation~\cite{thomas2005electrocardiogram,bashan2012network}. In clinical practice, this brain--body coupling is captured by polysomnography (PSG), which records heterogeneous signals spanning EEG, EOG, EMG, ECG and respiratory channels. These signals are informative not only for sleep staging, but also for assessing multi-system health risks that manifest through altered sleep physiology~\cite{lyu2025dynamic,abdalbari2022brain,thapa2024sleepfm,Fox2025AFT,goldammer2022investigation}.
Recent self-supervised foundation models for neural data, have shown that large-scale pretraining can learn transferable brain representations and reduce reliance on costly expert labels~\cite{jiang2024large,wang2026omni,chien2022maeeg,wang2026flexibrain,xia2026brain,wang2025slim,xia2026brainworld}. This paradigm has also been adopted for sleep analysis through large-scale pretraining on polysomnographic recordings~\cite{gerardy2023approach,thapa2024sleepfm}.
However, existing multimodal methods often align heterogeneous physiological signals in a unified representation space, overlooking their physiological organization~\cite{liu2024automatic,thapa2024sleepfm,phan2021xsleepnet,seo2020intra}. This flat fusion neglects a key property of PSG: CNS-derived signals and ANS-related signals follow distinct physiological manifolds, yet exhibit stage-dependent and time-varying synchronization~\cite{yetton2018quantifying,de2018dynamic}. In real-world deployments, these limitations are amplified under domain shifts and missing modalities.~\cite{chien2022maeeg}.

\begin{figure}[t]
    \centering
    \includegraphics[width=\linewidth]{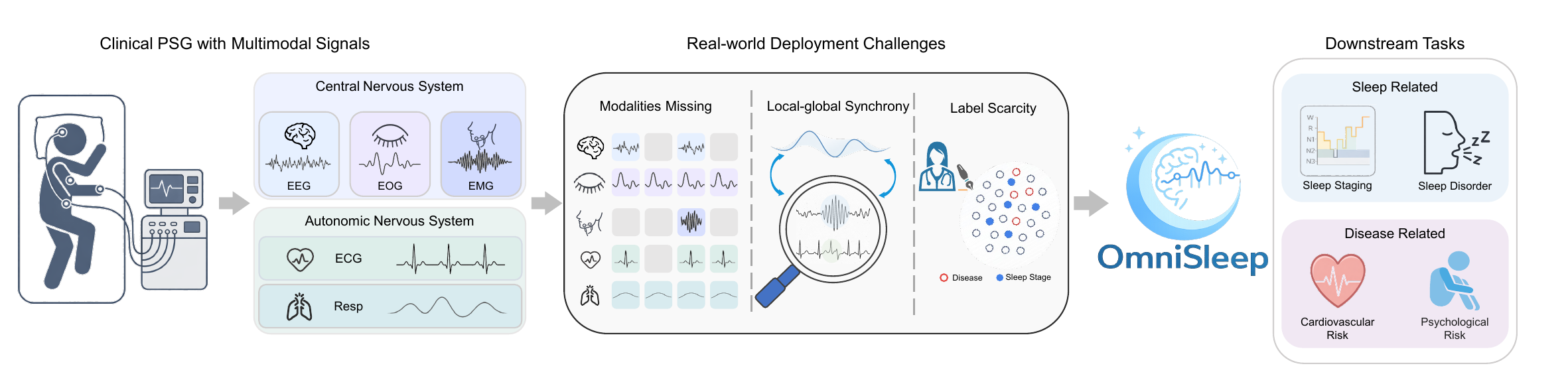}
    \caption{{\textbf{Omni-Sleep targets real-world PSG deployment.}
    }}
    \label{fig:fig1}
\end{figure}

To bridge this gap, we propose Omni-Sleep, a sleep foundation model that uses the CNS/ANS partition as a physiological prior for topology-constrained sleep representation learning. Pretrained on over 100,000 hours of multi-center PSG data, Omni-Sleep combines hierarchical contrastive learning to capture intra-system consistency and inter-system coupling with long-horizon latent prediction to model macro-scale sleep dynamics. Across multiple independent cohorts, Omni-Sleep demonstrates strong generalization and robustness in sleep staging and sleep-disorder assessment under label scarcity and missing-modality settings.
 The main contributions of this work are summarized as follows:
\begin{itemize}
    \item \textbf{Structure-Aware Foundation Model:} We propose Omni-Sleep, a sleep foundation model that uses the CNS/ANS split as a physiological prior while learning subsystem-level and cross-system representations.
    \item \textbf{Hierarchical Pretraining framework:} Omni-Sleep combines intra-system contrastive learning, inter-system alignment, and long-horizon latent prediction to capture structured CNS--ANS sleep dynamics.
    \item \textbf{SOTA Performance across Tasks:} Across datasets and modality-ablation settings, Omni-Sleep consistently improves out-of-domain transfer, label efficiency, and robustness to missing modalities, outperforming state-of-the-art baselines in sleep staging and multi-disease classification.
\end{itemize}

\section{Related Work}

\subsection{CNS-ANS Coupling in Sleep}
Sleep physiology reflects coordinated CNS-ANS dynamics, exhibiting stage-dependent and time-varying coupling between cortical activity and cardio-autonomic rhythms~\cite{thomas2005electrocardiogram,bashan2012network,lyu2025dynamic}. Such coupling is closely linked to sleep-stage transitions through systematic changes in heart rate variability and other autonomic markers, and shows structured temporal organization beyond static correlation~\cite{de2018dynamic}. Recent work further models brain--heart interactions by jointly analyzing EEG and cardiac dynamics across sleep~\cite{abdalbari2022brain}. These findings motivate structure-aware learning objectives that distinguish stable within-group patterns from cross-group synchronization, rather than relying on a single flat fusion operator.

\subsection{Multimodal PSG Representation Learning}
Multimodal PSG models typically improve performance via architectural fusion~\cite{yue2024research}, including early/late fusion~\cite{duan2021novel}, cross-attention~\cite{mostafaei2024novel}, modality-specific encoders with shared heads~\cite{kontras2024core}, and task-specific encoders with hand-crafted feature fusion~\cite{phan2021xsleepnet,seo2020intra}. While combining neural and cardio-respiratory signals benefits sleep staging and related tasks, most approaches still treat modalities as exchangeable views and align them in a unified representation space via generic attention or feature concatenation~\cite{liu2024automatic,thapa2024sleepfm}, overlooking their physiological organization. This motivates moving from flat fusion toward topology-aware objectives that respect structured cross-signal dependencies grounded in CNS--ANS physiology~\cite{de2018dynamic}.

\subsection{Limited Supervision and Partial Observability}
Real-world sleep monitoring is constrained by scarce expert labels, incomplete acquisition due to sensor dropouts and heterogeneous protocols, and distribution shifts across clinical sites and recording equipment~\cite{gerardy2023approach}. Inspired by recent self-supervised foundation models that learn transferable neural representations from large-scale brain recordings~\cite{jiang2024large,wang2026omni,xia2026brain}, sleep-specific SSL methods reduce label dependence via masked modeling and contrastive cross-modal alignment~\cite{chien2022maeeg,thapa2024sleepfm}, and improve robustness to missing modalities through imputation, distillation, or subset-based fusion~\cite{kontras2024core}. However, many pipelines still emphasize waveform reconstruction or flat global alignment in a single shared embedding space, which can blur complementary cues and degrade under partial observability and domain shift. We instead adopt a structured contrastive objective that enforces consistency within related signal groups while aligning across groups, yielding a representation that is more robust under real-world deployment conditions.

\subsection{Local--Global Synchrony in Sleep Dynamics}
Sleep modeling requires capturing both within-epoch micro-structures (e.g., spindles and K-complexes) and night-level, non-stationary dynamics~\cite{yetton2018quantifying}. Many pipelines process 30-second epochs and inject longer context using a downstream temporal module~\cite{seo2020intra}, typically an RNN/LSTM or, more recently, a Transformer~\cite{phan2022sleeptransformer,thapa2024sleepfm}. JEPA-style learning instead emphasizes latent-space prediction, prioritizing structured temporal regularities over waveform reconstruction for long-horizon dynamics~\cite{lecun2022path,assran2023self}. Motivated by this, we design complementary micro- and macro-scale pretraining pathways and enforce their consistency to induce local--global synchrony in the representation space.

\section{Methodology}
\label{sec:method}

\begin{figure}[t]
\centering
\includegraphics[width=\textwidth]{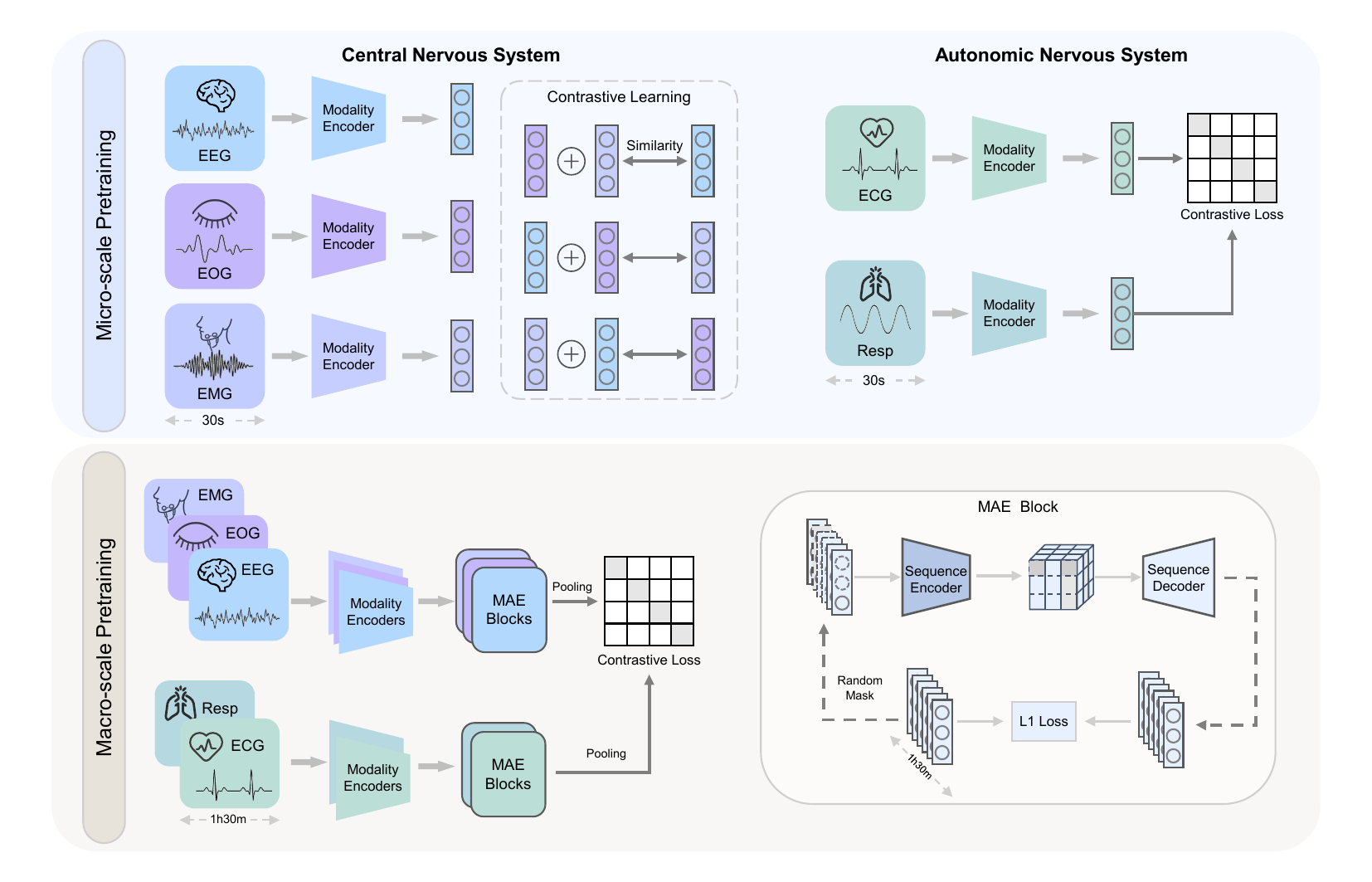}
\caption{\textbf{Omni-Sleep pretraining}.
\textbf{Micro-level} objectives capture intra-system consistency within CNS and ANS signals, while
\textbf{macro-level} objectives model inter-system synchronization and long-horizon sleep dynamics.}
\label{fig:fig2}
\end{figure}

\subsection{Problem Formulation}
PSG comprises a set of synchronized physiological modalities $\mathbf{X}=\{X^{(m)}\}_{m\in\mathcal{M}}$, where each $X^{(m)}$ is segmented into a sequence of 30\,s epochs. We explicitly encode the physiological information $\mathcal{M}$ by partitioning modalities into CNS subset $\mathcal{M}_C$ (EEG, EOG and EMG) and ANS subset $\mathcal{M}_A$ (ECG and respiration) with $\mathcal{M}=\mathcal{M}_C\cup\mathcal{M}_A$.
Our goal is to pretrain a multimodal encoder that preserves both within-group consistency and cross-group synchrony.

\subsection{Modality encoders}
For each modality $m$, we transform the raw epoch signal $X^{(m)}$ into a sequence of latent tokens
using a lightweight 1D convolutional patch encoder:
\begin{equation}
\mathbf{H}^{(m)} = f^{(m)}(\mathbf{X}^{(m)}), \qquad \mathbf{H}^{(m)}\in\mathbb{R}^{T\times D},
\end{equation}
where $T$ is the token length and $D$ is the latent dimension.

To model long-range temporal dependencies, tokens are passed through a modality-specific sequence encoder $g^{(m)}_{\text{seq}}$ (RoFormer with rotary positional encoding), providing contextualized tokens
$\mathbf{S}^{(m)} = g^{(m)}_{\text{seq}}(\mathbf{H}^{(m)})$, $\mathbf{S}^{(m)}\in\mathbb{R}^{T\times D}$
and an epoch-level embedding obtained by pooling:
\begin{equation}
\mathbf{s}^{(m)} = \text{Pool}\!\left(S^{(m)}\right)\in\mathbb{R}^{D}.
\end{equation}

We denote the projected representation for contrastive learning as
$\mathbf{z}^{(m)}=\psi(\mathbf{s}^{(m)})$, where $\psi(\cdot)$ is a small MLP projection head.

\subsection{Hierarchical Contrastive Learning}

Omni-Sleep optimizes a hierarchical contrastive objective that (i) enforces \textbf{intra-system consistency}
within CNS and within ANS (micro-scale), and (ii) preserves \textbf{inter-system coupling}
between pooled CNS and pooled ANS summaries (macro-scale).
\paragraph{InfoNCE objective.}
Given an anchor representation $\mathbf{z}_i$ and its positive $\mathbf{z}_i^+$, we use standard InfoNCE:
\begin{equation}
\ell_{\text{NCE}}(\mathbf{z}_i,\mathbf{z}_i^+) =
-\log \frac{\exp(\text{sim}(\mathbf{z}_i,\mathbf{z}_i^+)/\tau)}
{\sum_{j=1}^{B} \exp(\text{sim}(\mathbf{z}_i,\mathbf{z}_j)/\tau)},
\label{eq:nce}
\end{equation}
where $\text{sim}(\cdot,\cdot)$ is cosine similarity, $\tau$ is a temperature, $B$ is the batch size,
and negatives are other samples in the mini-batch.

\paragraph{Micro-scale (intra-system) contrastive learning.}
For each modality $m$ within a subsystem $\mathcal{S}\in\{\mathcal{M}_C,\mathcal{M}_A\}$,
we construct the positive as the mean of the other modalities in the same subsystem and same epoch:
\begin{equation}
\mathbf{z}^{(m)+} = \frac{1}{|\mathcal{S}|-1}\sum_{m'\in\mathcal{S}\setminus\{m\}} \mathbf{z}^{(m')}.
\label{eq:intra_pos}
\end{equation}
The intra-system loss is
\begin{equation}
\mathcal{L}_{\text{intra}} = \sum_{\mathcal{S}\in\{\mathcal{M}_C,\mathcal{M}_A\}}
\sum_{m\in\mathcal{S}} \ell_{\text{NCE}}\!\left(\mathbf{z}^{(m)},\mathbf{z}^{(m)+}\right).
\label{eq:lintra}
\end{equation}
This encourages each modality to agree with the shared subsystem representation, improving robustness when some channels are absent.

\paragraph{Macro-scale (inter-system) alignment.}
To explicitly model brain--body coupling, we form pooled subsystem summaries
\begin{equation}
\mathbf{z}_C = \frac{1}{|\mathcal{M}_C|}\sum_{m\in\mathcal{M}_C}\mathbf{z}^{(m)},\qquad
\mathbf{z}_A = \frac{1}{|\mathcal{M}_A|}\sum_{m\in\mathcal{M}_A}\mathbf{z}^{(m)}.
\end{equation}
We then align CNS and ANS summaries using a symmetric InfoNCE:
\begin{equation}
\mathcal{L}_{\text{inter}} =
\ell_{\text{NCE}}(\mathbf{z}_C,\mathbf{z}_A) + \ell_{\text{NCE}}(\mathbf{z}_A,\mathbf{z}_C).
\label{eq:linter}
\end{equation}

\paragraph{Contrastive objective.}
The topology-aware contrastive loss is
\begin{equation}
\mathcal{L}_{c} = \mathcal{L}_{\text{intra}} + \lambda\,\mathcal{L}_{\text{inter}},
\label{eq:lc}
\end{equation}
where $\lambda$ controls the strength of CNS--ANS coupling.

\subsection{Long-horizon latent masked modeling (macro-scale)}
Local morphology within a single epoch is insufficient to capture long-range sleep dynamics.
We therefore introduce a long-horizon masked prediction objective over windows comprising multiple consecutive epochs
(e.g., an extended context such as 1.5\,h). For each modality $m$, we collect the epoch embeddings in a window:
\begin{equation}
\mathbf{E}^{(m)} = [\mathbf{h}^{(m)}_1,\ldots,\mathbf{h}^{(m)}_L]\in\mathbb{R}^{L\times D},
\end{equation}
where $L$ is the number of epochs in the long window.
We randomly mask a subset of epoch positions $\Omega$ and feed the unmasked sequence to a temporal RoFormer
to predict embeddings at masked positions:
\begin{equation}
\hat{\mathbf{h}}^{(m)}_t = g_{\text{pred}}^{(m)}\!\left(\mathbf{E}^{(m)}_{\setminus \Omega}\right),\quad t\in\Omega.
\end{equation}
We use an $\ell_1$ reconstruction loss to encourage robust, long-horizon representations:
\begin{equation}
\mathcal{L}_{p} = \sum_{m\in\mathcal{M}}\sum_{t\in\Omega} \left\lVert \mathbf{h}^{(m)}_t - \hat{\mathbf{h}}^{(m)}_t \right\rVert_{1}.
\label{eq:lp}
\end{equation}
This objective complements $\mathcal{L}_c$ by enforcing consistency over extended temporal context, improving transfer under domain shift and incomplete observations.

\subsection{Overall training objective and schedule}
The final pretraining loss is
\begin{equation}
\mathcal{L} = \alpha\,\mathcal{L}_{c} + \beta\,\mathcal{L}_{p},
\label{eq:total}
\end{equation}
with weights $\alpha$ and $\beta$.
We use a two-stage schedule: (i) warm-up pretraining with topology-aware contrastive learning ($\mathcal{L}_c$),
followed by (ii) joint optimization of $\mathcal{L}_c$ and long-horizon masked prediction ($\mathcal{L}_p$).

\subsection{Implementation details}
Omni-Sleep contains about \textbf{56 M parameters} in total.
Pretraining is conducted on a single NVIDIA A800 GPU for 40 epochs, with $\sim$40 minutes per epoch.
We optimize using AdamW with a cosine learning-rate schedule.

\section{Experiments}
\label{sec:experiments}

To evaluate the efficacy of Omni-Sleep in learning unified and robust representations, we designed an experimental protocol spanning multiple scales of data availability, domain shifts, and clinical pathologies.

\subsection{Experimental Setup}
\label{subsec:exp_setup}
\textbf{Datasets.} Our pre-training corpus comprises over 100,000 hours of multi-center PSG recordings from SHHS \cite{quan1997sleep}, WSC \cite{young2009rationale}, and MESA \cite{chen2015racial}. A random subset of 1,159 SHHS subjects was strictly held out for downstream disease classification. For independent out-of-domain (OOD) evaluation, we utilized ISRUC-Sleep Subgroup I (100 subjects)\cite{khalighi2016isruc} and CinC (994 subjects) \cite{ghassemi2018you}, ensuring absolute separation from the pre-training phase. All PSG recordings underwent subject-specific normalization. To efficiently standardize multimodal inputs, high-frequency channels (EEG, EOG, EMG, ECG) were uniformly resampled to 100 Hz, while lower-frequency respiratory and airflow channels were resampled to 10 Hz.

\noindent\textbf{Implementation Details.} Omni-Sleep is implemented in PyTorch and optimized with AdamW using a cosine learning rate schedule. We compare Omni-Sleep (56M parameters) with SleepFM (4.4M) and SleepGPT (134M) as baselines. Representations are evaluated using both linear probing with a frozen backbone and full fine-tuning in few-shot settings. Performance is measured by Macro F1-score, Cohen's Kappa ($\kappa$), and Accuracy for sleep staging, and by AUROC for disease classification.

\begin{table}[htbp]
\centering
\footnotesize
\caption{\textbf{Linear-probing performance on sleep staging across PSG channel sets.}
We evaluate frozen representations using different channel configurations. \textbf{Omni-Sleep (ours)} denotes the final proposed strategy. CNS and ANS refer to central and autonomic nervous system signal sets, respectively. Best results are highlighted in \textbf{bold}.}

\label{tab:performance_comparison}
\setlength{\tabcolsep}{3.5pt}
\resizebox{\linewidth}{!}{%
\begin{tabular}{llcccccccc}
    \toprule
    \multirow{2}{*}{\textbf{Modality}} & \multirow{2}{*}{\textbf{Model}} & \multicolumn{3}{c}{\textbf{Overall ($\uparrow$)}} & \multicolumn{5}{c}{\textbf{Class-wise F1 ($\uparrow$)}} \\
    \cmidrule(lr){3-5} \cmidrule(lr){6-10}
     & & Acc. & $\kappa$ & MF1 & W & N1 & N2 & N3 & REM \\
    \midrule

    \multirow{3}{*}{\textbf{EEG}}
    & SleepFM \cite{thapa2024sleepfm}  & 65.8 & 0.54 & 64.3 & 76.1 & 37.8 & 74.2 & 73.4 & 59.9 \\
    & SleepGPT \cite{Huang2026AUT} & 68.6 & 0.59 & 67.2 & 79.7 & 45.1 & 72.3 & 75.4 & 63.6 \\
    & Omni-Sleep (ours) & \textbf{74.0} & \textbf{0.66} & \textbf{73.5} & \textbf{82.7} & \textbf{56.4} & \textbf{75.7} & \textbf{77.1} & \textbf{75.6} \\

    \midrule

    \multirow{3}{*}{\textbf{CNS}}
    & SleepFM \cite{thapa2024sleepfm}  & 70.5 & 0.61 & 69.1 & 78.8 & 42.6 & 78.9 & 74.1 & 71.1 \\
    & SleepGPT \cite{Huang2026AUT}& 70.8 & 0.62 & 69.9 & 80.3 & 48.8 & 73.9 & 76.3 & 70.3 \\
    & Omni-Sleep (ours) & \textbf{76.7} & \textbf{0.70} & \textbf{76.2} & \textbf{84.2} & \textbf{59.6} & \textbf{78.4} & \textbf{78.5} & \textbf{80.3} \\

    \midrule

    \multirow{2}{*}{\textbf{ANS}}
    & SleepFM \cite{thapa2024sleepfm}  & 58.4 & 0.42 & 57.1 & 62.5 & 30.2 & 69.8 & 71.5 & 51.5 \\
    & Omni-Sleep (ours) & \textbf{66.3} & \textbf{0.56} & \textbf{65.3} & \textbf{71.5} & \textbf{44.8} & \textbf{69.7} & \textbf{74.7} & \textbf{66.1} \\

    \midrule

    \multirow{2}{*}{\textbf{Full}}
    & SleepFM \cite{thapa2024sleepfm}  & 71.8 & 0.63 & 70.0 & 79.0 & 43.9 & 79.3 & 74.3 & 71.7 \\
    & SleepGPT \cite{Huang2026AUT}  & 70.8 & 0.62 & 69.9 & 80.3 & 48.8 & 73.9 & 76.3 & 70.3 \\
    & Omni-Sleep (ours) & \textbf{77.8} & \textbf{0.71} & \textbf{77.3} & \textbf{85.1} & \textbf{60.6} & \textbf{79.5} & \textbf{78.9} & \textbf{82.4} \\
    \bottomrule
\end{tabular}%
}
\end{table}

\subsection{Out-of-Domain Generalization}
\label{subsec:ood_generalization}
To assess intrinsic feature quality and transferability, we evaluated Omni-Sleep on the external CinC 2018 dataset under strict linear probing. By explicitly modeling the distinct manifolds of the CNS and ANS during pre-training, Omni-Sleep extracts invariant features that generalize to unseen clinical environments, outperforming existing flat-fusion foundation models without task-specific weight updates. As detailed in Table \ref{tab:performance_comparison}, Omni-Sleep consistently establishes new state-of-the-art benchmarks across all evaluated channel combinations. In the Full modality setting, our model achieves an Accuracy of 77.8\% and a Macro F1 of 77.3\%, yielding substantial absolute improvements over existing models. Crucially, Omni-Sleep demonstrates remarkable resilience to sensor dropouts; even in restricted scenarios such as ANS-only or EEG-only, it substantially exceeds the performance of both SleepFM and SleepGPT. This validates that our hierarchical alignment strategy effectively extracts generalized and robust representations independent of sensor availability.

\begin{figure}[htbp]
    \centering
    \includegraphics[width=1\columnwidth]{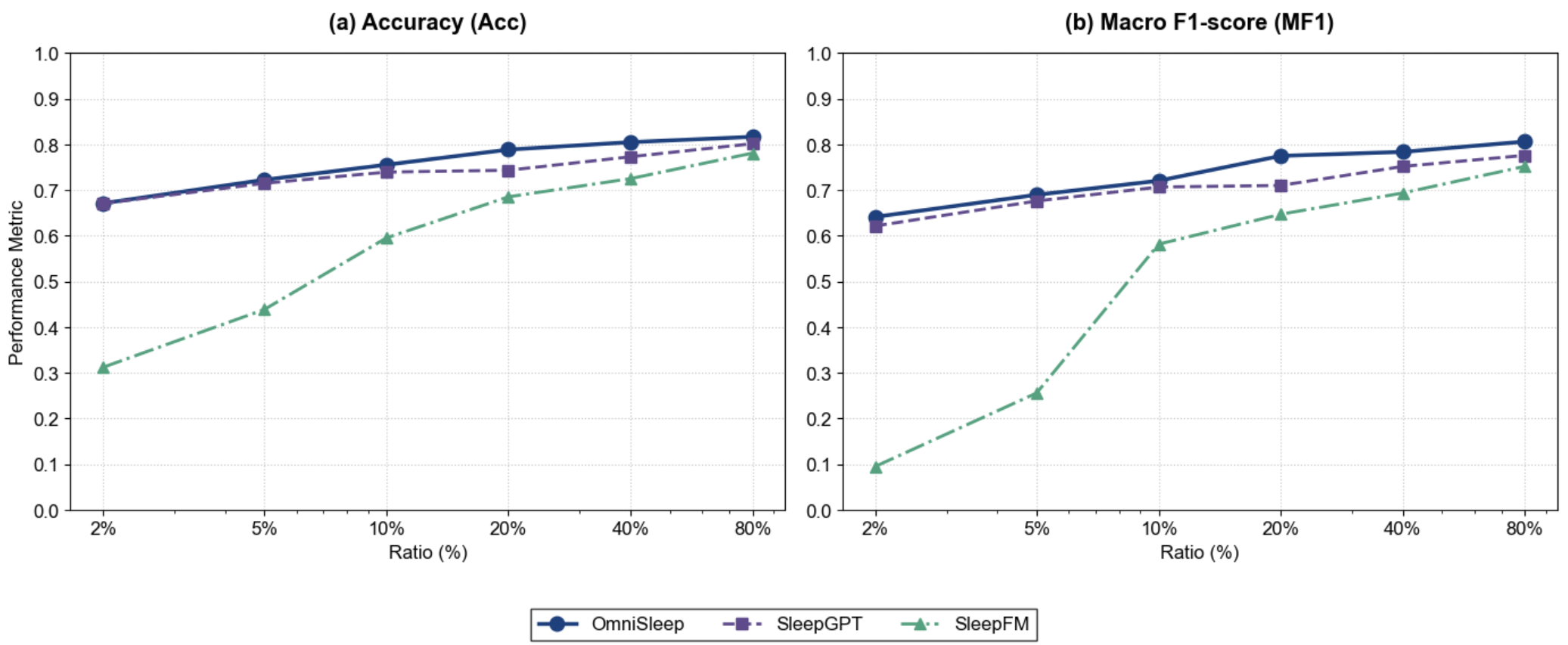}
    \caption{Few-shot sleep staging on ISRUC-I.}
    \label{fig:fewshot_performance}
\end{figure}

\subsection{Label Efficiency in Clinical Scenarios}
\label{subsec:label_efficiency}

To evaluate label efficiency under realistic low-annotation settings, we fine-tuned the model using 2\%--80\% labeled data.
As shown in Fig.~\ref{fig:fewshot_performance}, Omni-Sleep consistently outperformed SleepGPT and SleepFM across all ratios, remaining competitive even at 2\% supervision and progressively approaching the fully supervised regime as label availability increased. These findings indicate that the proposed temporal predictive pre-training effectively captures clinically meaningful transition dynamics, thereby reducing downstream annotation requirements.

\subsection{Multi-Label Disease Classification}
\label{subsec:pathology_detection}

We evaluated Omni-Sleep for multi-disease classification on SHHS1 across sleep-related, psychiatric, respiratory, and cardiovascular conditions. Binary labels were defined using established clinical criteria, including AHI $>5$ for OSA, ESS $\geq 12$ for hypersomnia, MCS $\leq 42$ for clinical depression, and ``So Blue'' score $\geq 3$ for depressive affect. Respiratory outcomes, including COPD and emphysema, and cardiovascular outcomes, including heart failure and stroke, were derived from physician-confirmed diagnoses. As shown in \cref{tab:disease_comparison_combined}, linear probing on frozen Omni-Sleep representations consistently outperforms demographic baselines and prior foundation models. The ablation variant Omni-Sleep$^\ast$ further suggests that long-horizon temporal modeling contributes to classification of respiratory and cardiovascular conditions, while modality-aggregation results demonstrate robustness to partial observability and sensor dropouts.

\begin{table}[htbp]
    \centering
    \caption{\textbf{Performance comparisons on multi-label disease classification.} We report AUROC across disease classification tasks via linear probing. Values are reported as mean $\pm$ standard deviation over five runs, with the best results highlighted in \textbf{bold}. Omni-Sleep$^\ast$ denotes an ablation variant without the RoFormer module.}

    \label{tab:disease_comparison_combined}
    \setlength{\tabcolsep}{4pt}
    \footnotesize
    \resizebox{\linewidth}{!}{%
    \begin{tabular}{@{} l cc cc @{}}
        \toprule
        \multirow{2}{*}{Method} & \multicolumn{2}{c}{Sleep Disorders} & \multicolumn{2}{c}{Respiratory Disorders} \\
        \cmidrule(lr){2-3} \cmidrule(lr){4-5}
        & OSA & Hypersomnia  & COPD & Emphysema \\
        \midrule
        Demographic \cite{Thapa2026AMS}
        & 0.774$\pm$0.017 & 0.423$\pm$0.328 & 0.562$\pm$0.175 & 0.632$\pm$0.201 \\
        SleepGPT \cite{Huang2026AUT}
        & 0.597$\pm$0.029 & 0.460$\pm$0.123 & 0.626$\pm$0.119 & 0.649$\pm$0.071 \\
        Omni-Sleep$^\ast$
        & 0.789$\pm$0.044 & 0.471$\pm$0.053 & 0.665$\pm$0.191 & 0.692$\pm$0.060 \\
        Omni-Sleep (ours)
        & \textbf{0.841$\pm$0.012} & \textbf{0.551$\pm$0.109} & \textbf{0.750$\pm$0.158} & \textbf{0.750$\pm$0.043} \\

        \midrule
        \midrule
        \multirow{2}{*}{Method} & \multicolumn{2}{c}{Psychological} & \multicolumn{2}{c}{Cardiovascular} \\
        \cmidrule(lr){2-3} \cmidrule(lr){4-5}
        & Clin. Depression & Depressive Affect & Heart Fail & Stroke \\
        \midrule
        Demographic \cite{Thapa2026AMS}
        & 0.513$\pm$0.037 & 0.545$\pm$0.124 & 0.625$\pm$0.024 & 0.707$\pm$0.144 \\
        SleepGPT \cite{Huang2026AUT}
        & 0.516$\pm$0.078 & 0.583$\pm$0.091 & 0.409$\pm$0.147 & 0.507$\pm$0.043 \\
        Omni-Sleep$^\ast$
        & \textbf{0.545$\pm$0.057} & 0.565$\pm$0.053 & 0.796$\pm$0.063 & 0.735$\pm$0.078 \\
        Omni-Sleep (ours)
        & 0.519$\pm$0.051 & \textbf{0.623$\pm$0.041} & \textbf{0.825$\pm$0.066} & \textbf{0.739$\pm$0.078} \\
        \bottomrule
    \end{tabular}%
    }
\end{table}

\section{Conclusion}
We propose Omni-Sleep, a novel multimodal foundation model that uses the CNS/ANS partition as a physiological prior for topology-constrained sleep representation learning. Pre-trained on over 100,000 hours of multimodal PSG data, Omni-Sleep combines intra-system consistency, inter-system synchronization, and long-horizon latent prediction to capture subsystem-level factors, CNS--ANS coupling, and macro-scale sleep dynamics. Across datasets and modality-ablation settings, Omni-Sleep improves generalization, few-shot label efficiency, and missing-modality robustness for sleep staging and sleep-related disorder classification, highlighting the value of physiological hierarchy for sleep representation learning.

\section*{Acknowledgements}
This work was supported in part by the National Natural Science Foundation of China, under Grant Nos. 3254100307 and 62472206; the Brain Science and Brain-like Intelligence Technology National Science and Technology Major Project, under Grant No. 2021ZD0200500; the National Key R\&D Program of China, under Grant No. 2025YFC3410000; and the Shenzhen Science and Technology Innovation Committee, under Grant Nos. RCYX20231211090405003 and JCYJ20220818100213029. Additional support was provided by the Guangdong Basic and Applied Basic Research Foundation, under Grant No. 2026B1515020099; the Guangdong S\&T Program, under Grant No. 2026B0101110003; the Shanghai Municipal Special Program for Basic Research on General AI Foundation Models, under Grant No. 2025SHZDZX026D05; the Shenzhen Loop Area Institute, under Grant No. FPF10120250012; the open research fund of the Guangdong Provincial Key Laboratory of Mathematical and Neural Dynamical Systems; the Center for Computational Science and Engineering at Southern University of Science and Technology; and the Shenzhen Key Laboratory of Smart Healthcare Engineering.

\section*{Competing Interests}
The authors have no competing interests to declare that are relevant to the content of this article.

\bibliography{ref}
\bibliographystyle{plain}

\end{document}